\documentclass[letterpaper, 10pt, conference]{ieeeconf}

\IEEEoverridecommandlockouts
\usepackage{graphicx}
\usepackage{amsmath,amssymb}
\usepackage{array}
\usepackage{booktabs}
\usepackage{multirow}
\usepackage{pifont}
\usepackage{xcolor}
\usepackage{cite}
\usepackage{caption}
\usepackage{subcaption}
\usepackage[colorlinks=true,
            allcolors=blue,
            breaklinks=true,
            bookmarks=false,
            pdftitle={DriveReferee: Geometric Safety Verdicts Need Not Be Learned for Driving World-Action Models},
            pdfauthor={Fengcheng Yu, Dhruv Parikh, Junjie Ye, Maulik Bhatt, Thang Vu, Igor Vasiljevic, Vitor Guizilini, Yue Wang},
            pdfsubject={Autonomous driving, world-action models, trajectory verification},
            pdfkeywords={autonomous vehicle navigation, integrated planning and learning, robot safety}
           ]{hyperref}
\usepackage{cleveref}
\usepackage{makecell}
\crefname{section}{Sec.}{Secs.}
\crefname{subsection}{Sec.}{Secs.}
\crefname{figure}{Fig.}{Figs.}
\crefname{equation}{Eq.}{Eqs.}
\crefname{table}{Table}{Tables}
\graphicspath{{assets/}}

\newcommand{\method}{DriveReferee}

\newcommand{\traj}{\tau}
\newcommand{\map}{M}
\newcommand{\referee}{R(\traj,\map)}

\title{\LARGE \bf \method{}:
Geometric Safety Verdicts Need Not \\ Be Learned for Driving World-Action Models}

\author{Fengcheng Yu$^{1}$, Dhruv Parikh$^{1}$, Junjie Ye$^{1}$, Maulik Bhatt$^{2}$,\\
Thang Vu$^{2}$, Igor Vasiljevic$^{3}$, Vitor Guizilini$^{3\dagger}$, Yue Wang$^{1\dagger}$\\[4pt]
$^{1}$University of Southern California \quad $^{2}$Woven by Toyota \quad $^{3}$Toyota Research Institute\\[2pt]
$^{\dagger}$Equal advising}

\makeatletter
\def\IEEEaftertitletext#1{\def\@IEEEaftertitletext{#1}}
\makeatother

\begin{document}

\maketitle
\thispagestyle{empty}
\pagestyle{empty}

\begin{abstract}
Generative world-action models (WAMs) jointly generate future video and vehicle actions, while their action branches remain primarily optimized by expert imitation. 
Yet imitation provides no explicit closed-loop geometric verdict for generated trajectories, making verification important during both training and deployment.
Closed-loop evaluators can check collision and drivable-area violations, but require privileged scene state unavailable at deployment.
Existing approaches often address this gap by learning a verifier from sensor features. For these geometric checks, the rule itself is explicit. For example, collision is determined by whether the rolled-out ego footprint overlaps occupied vehicle space. What is unavailable at deployment is the scene state needed to apply the rule.
We introduce \method{}, which uses a learned geometry readout to predict the scene representation from camera observations and executes the geometric safety rule directly rather than learning it. 
The resulting analytic referee evaluates collision and drivable-area safety from a scene state and candidate trajectory. During training, it scores self-sampled trajectories on ground-truth state and distills the resulting preferences into the WAM policy. At deployment, the same referee evaluates generated trajectories on this predicted state and selects a safer alternative when needed. 
The analytic referee requires no verdict-specific training, and its decisions
follow an explicit geometric rule. Under matched candidates and inference
budgets, it matches or outperforms all learned-verifier and heuristic baselines.
Given the same predicted state and trajectory, learning the verdict provides
no measurable downstream gain despite requiring tens of thousands of
evaluator-labeled training examples.
On the full NAVSIM navtest, \method{} reaches 92.02 PDMS with single-camera visual input and no external training data.
\end{abstract}

\section{INTRODUCTION}
\label{sec:intro}

\begin{figure}[t]
\centering
\begin{subfigure}[t]{0.385\columnwidth}
  \centering
  \includegraphics[width=\linewidth]{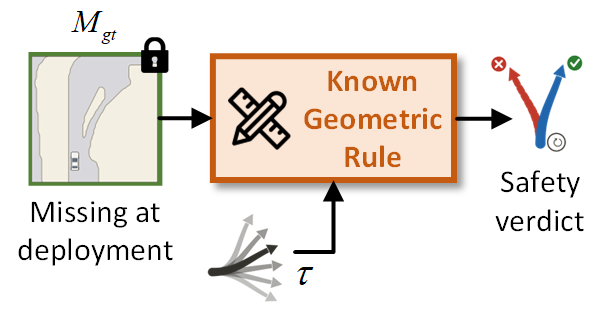}
  \caption{Privileged evaluation: the rule is known, but $\map_{\mathrm{gt}}$
  is unavailable at deployment.}
  \label{fig:teaser-a}
\end{subfigure}\hfill
\begin{subfigure}[t]{0.595\columnwidth}
  \centering
  \includegraphics[width=\linewidth]{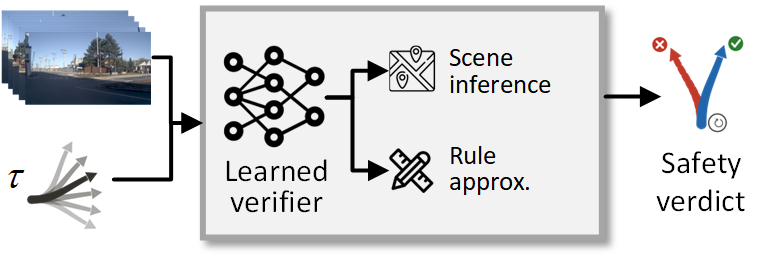}
  \caption{Learned verifier: one network must infer the scene and approximate
  the rule at once.}
  \label{fig:teaser-b}
\end{subfigure}\\[3pt]
\begin{subfigure}[t]{\columnwidth}
  \centering
  \includegraphics[width=\linewidth]{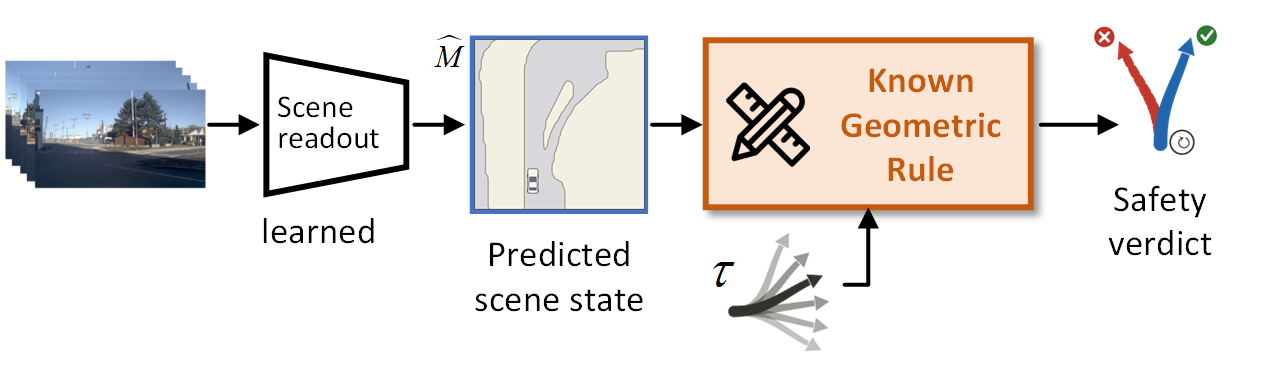}
  \caption{\method{}: predict the missing state $\hat\map$ and execute the
  known rule.}
  \label{fig:teaser-c}
\end{subfigure}
\caption{\textbf{\method{} recovers the state, instead of the rule.} The
geometric safety verdict is a computation whose input, not whose logic, is
missing at deployment.}
\label{fig:teaser}
\vspace{-20pt}
\end{figure}

Generative world-action models (WAMs) jointly generate future video and vehicle actions, coupling scene evolution with motion
decisions~\cite{liu2026universeunifiedvideoaction,liu2026driveva,shi2026drivewam,zhou2026drivedreamer,li2026metis}.
As WAMs can produce multiple plausible action futures, deciding whether a generated trajectory is safe becomes as important as generating it.
For geometric failures such as collision and departure from the drivable area, however, the safety rule itself is not unknown: closed-loop evaluators already determine these outcomes from the scene geometry and the candidate trajectory~\cite{dauner2024navsim}. 
However, the obstacle to applying the same logic during the deployment phase is that the evaluator relies on privileged scenario states that are inaccessible via onboard observations. In other words, \emph{what is missing at deployment is the evaluator's input, not its geometric rules} (\cref{fig:teaser}).

Existing approaches often bridge this gap by learning a verifier that predicts safety scores or rankings directly from sensor features and candidate trajectories~\cite{li2024hydra,li2025hydra,li2025end,wu2026distill}. This combines two distinct tasks in a single learned mapping: inferring scene state from observations and computing a geometric safety verdict for a candidate trajectory.
Scene inference must be learned from visual observations, while these geometric safety verdicts can be computed directly from scene geometry, the vehicle footprint, and the candidate trajectory.
When the geometric rule is already known, it can instead be applied
directly once the required scene state is available. This avoids
verdict-specific training and keeps the final safety decision explicit.
We therefore ask whether directly executing the known rule can provide
the same or better downstream value than learning the verdict from data.

We therefore introduce \method{}, a referee-guided world-action modeling framework that integrates geometric safety reasoning into both policy learning and test-time generation. 
Instead of appending a separate verifier after the planning stage, \method{} uses the WAM's own random action generation mechanism to establish a feedback loop between generation and verification. 
During training, the policy self-samples candidate trajectories. The analytic
referee evaluates them using privileged ground-truth scene state to form
preference pairs, which are then used to train the WAM. Separately, we train
a geometry readout to predict the drivable-area and vehicle-occupancy maps
from camera features, using ground-truth scene maps as supervision.
At deployment, only the predicted maps are available. The same analytic
referee uses these maps to check the default plan and, when needed, compare
additional samples and select a safer plan. In our setting, the missing state
is the drivable-area and vehicle-occupancy maps needed by the referee. We
learn to predict this state from camera observations while keeping the
geometric safety rule fixed.

We first compare the analytic referee with the corresponding NAVSIM safety checks~\cite{dauner2024navsim}. We then measure how its predictions change when ground-truth scene state is replaced by the self-predicted map. 
We then compare alternative verdict sources under matched candidates, decision protocols, and inference budgets. We also compare the analytic referee with a learned verifier that receives the same predicted map and candidate trajectory. Even with 66,385 evaluator-labeled training examples, the learned verifier provides no measurable improvement over the analytic rule. With the referee integrated into training and deployment, DriveReferee reaches 92.02/91.56 PDMS/EPDMS on the full NAVSIM \texttt{navtest}~\cite{dauner2024navsim} while preserving video-generation quality.

Our main contributions are:
\begin{itemize}
    \item We introduce a verification decomposition for driving WAMs that separates learned scene-state inference from analytic geometric verdict computation, enabling the known safety rule to be executed directly on
    predicted state.

    \item We propose DriveReferee, a referee-guided WAM framework that reuses the same zero-parameter analytic referee in two complementary placements: on privileged ground-truth state for training-time preference distillation, and on self-predicted state for alarm-gated candidate selection at deployment.

    \item We isolate the value of learning the final verdict through controlled matched-budget comparisons and a same-map learned replacement that removes scene-representation quality as a confounding factor.
\end{itemize}
\section{RELATED WORK}
\label{sec:related}

Among generative WAMs, which jointly model future video and driving actions~\cite{liu2026universeunifiedvideoaction,liu2026driveva,shi2026drivewam,zhou2026drivedreamer,li2026metis}, driving actions are still primarily learned through imitation of logged expert trajectories.
Verification has largely been studied through oracle best-of-$N$ evaluation rather than as a deployable mechanism. For example, Metis~\cite{li2026metis} follows the best-of-$N$ protocol of AutoVLA~\cite{zhou2026autovla}, using an oracle scorer to measure the headroom available from multiple samples. How to identify the safer candidate without privileged information therefore remains underexplored in this family.

The broader end-to-end planning literature incorporates safety supervision in several ways during training. Evaluator-derived preferences can guide policy optimization~\cite{shang2026drivedpo}, while synthesized hard negatives provide informative unsafe examples~\cite{wang2026beyond}. Other methods use privileged rule-based metrics as regression targets~\cite{liu2026perceptdrive} or distill them into learned trajectory scorers for later selection~\cite{li2024hydra}. At test time, learned evaluators have been used to score world-model rollouts~\cite{li2025end}, verify or refine predicted trajectories~\cite{he2026drivever}, steer trajectory
search~\cite{xu2026test,jiao2025evadrive}, and rank sampled
candidates~\cite{sun2025minddrive}. These learned evaluators predict metrics, confidence, or rankings from scene features and candidate trajectories; \method{} instead asks whether the final geometric safety verdict needs to be learned at all.

Other work applies explicit safety rules once the required scene information is available. Responsibility-Sensitive Safety (RSS) defines formal rules for safe vehicle interactions~\cite{shalev2017formal}. Safety filters and runtime-assurance systems use known safety constraints to monitor or correct learned policies~\cite{hsu2023safety,sha2001using}. PDM-Closed performs rule-based trajectory ranking using privileged scene information~\cite{dauner2023parting,dauner2024navsim}.
Recent CoPhy similarly derives physical safety objectives from a predicted semantic BEV representation~\cite{wu2026distill}. \method{} differs in three respects: it operates inside a WAM that supplies its own candidates, reuses the same referee for training-time preference supervision and deployment-time selection, and directly compares learned and analytic verdicts, including a same-map control in which both receive the same predicted map and candidate trajectory.
\section{METHOD}
\label{sec:method}

\begin{figure*}[t]
\centering
\includegraphics[width=1.8\columnwidth]{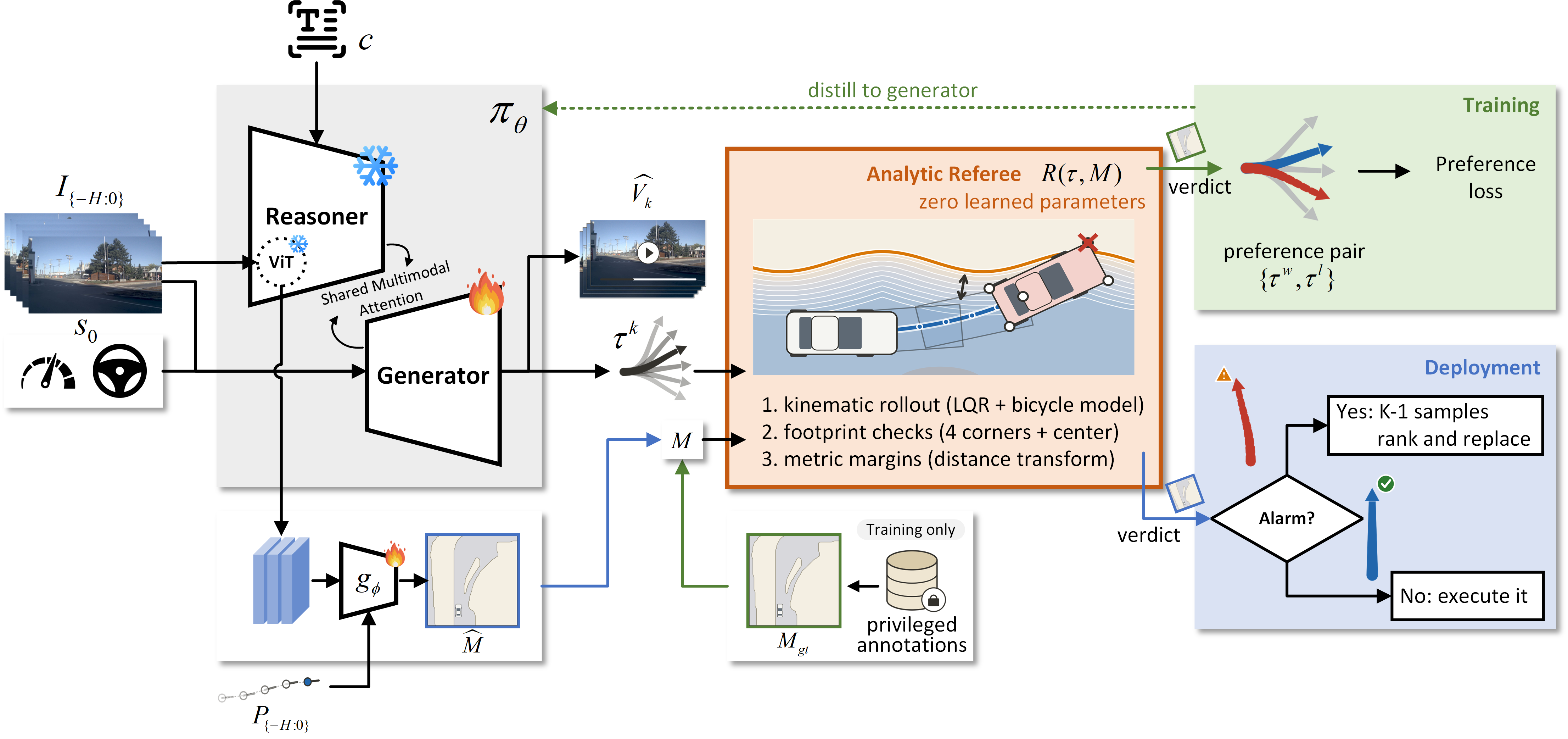}
\caption{\textbf{\method{} overview.} The policy $\pi_\theta$ (frozen
reasoner, trained generator) generates future video $\hat V_k$ and candidate plans $\tau_k$; the readout $g_\phi$ predicts the BEV scene state $\hat\map=(\hat M_{\mathrm{drv}},\hat M_{\mathrm{veh}})$, containing drivable-area and vehicle-occupancy maps, from frozen visual features and ego poses $P_{-H:0}$. The analytic referee $\referee$, with no learned parameters, is placed twice: at training (green) on $\map_{\mathrm{gt}}$ to form preference pairs $(\traj^w,\traj^l)$ distilled into the generator (\cref{eq:pref}); at deployment (blue) on $\hat\map$ to gate the default plan and, on alarm, rank $K{-}1$ further samples on the same map.}
\label{fig:method}
\vspace{-15pt}
\end{figure*}

\subsection{Problem Setup and Overview}
\label{sec:method-overview}

We index time relative to the current planning step. The $H+1$ observed steps are indexed by $i=-H,\ldots,0$, with $i=0$ the current step, and the $T_f$ future planning steps by $j=1,\ldots,T_f$.
Let $I_{-H:0}=(I_{-H},\ldots,I_0)$ denote the latest $H+1$
front-camera frames, $c$ the benchmark's discrete navigation command, and
$s_0=(v_0,\psi_0^{\mathrm{glob}})$ the current ego speed $v_0$ and
global-frame heading $\psi_0^{\mathrm{glob}}$ recorded in the log.
The policy observation is $o=(I_{-H:0},c,s_0)$.
A future plan is represented as
$\tau=\{(x_j,y_j,\psi_j)\}_{j=1}^{T_f}$, where
$(x_j,y_j,\psi_j)$ is the ego pose at future step $j$.
All future poses are expressed in the current ego frame, with origin at the rear axle and the $x$-axis pointing forward.
For the observed frames,
$P_{-H:0}=(P_{-H},\ldots,P_0)$ denotes the ego-pose history, where
$P_i=(x_i,y_i,\psi_i)$ is the ego pose at observed step $i$, expressed in the current ego frame; hence $P_0=(0,0,0)$.
The policy observes $o$, while the perception readout additionally
receives $P_{-H:0}$. For two plans, $d(\tau,\tau')$ denotes the mean
distance between corresponding waypoint positions.

DriveReferee contains three components (\cref{fig:method}): 
(1) a world-action policy $\pi_\theta$ that generates future video and candidate trajectories,
(2) a perception readout $g_\phi$ that predicts the scene state, and
(3) an analytic referee $\referee$ that evaluates candidate trajectories.

\noindent\textbf{World-action policy.}
The policy $\pi_\theta$ jointly generates future video and vehicle actions~\cite{nvidia2026cosmos3omnimodalworld}. It couples a frozen reasoner tower, which reads the navigation command, with a generation tower that produces the video and action tokens; $\theta$ denotes the parameters of the generation tower, the only ones we train. Each policy sample yields a future video $\hat V_k$ and a decoded plan $\traj_k$; repeated sampling therefore provides multiple candidate trajectories without an additional proposal network. We use $K_t$ self-samples for training-time preference construction and at most $K$ samples at deployment.

\noindent\textbf{Perception readout.}
The readout $g_\phi$ operates on frozen visual features from $I_{-H:0}$ and uses the associated ego-pose history $P_{-H:0}$ to predict the scene state required by the referee, which we represent as a BEV map $\map=(\map_{\mathrm{drv}},\map_{\mathrm{veh}})$, where
$\map_{\mathrm{drv}}$ is the static drivable-area map and
$\map_{\mathrm{veh}}$ is the time-indexed vehicle-occupancy map.
$\map_{\mathrm{gt}}$ denotes the ground-truth version of this
representation, obtained from privileged training annotations, while
$\hat\map$ denotes the version predicted by $g_\phi$ at deployment.

\noindent\textbf{Analytic referee.}
The analytic referee $\referee$ evaluates candidate plans using the scene
representation. The same referee is used in both placements: during training,
it evaluates self-sampled trajectories on $\map_{\mathrm{gt}}$ to form
preference pairs for policy training (\cref{sec:method-train}); at deployment,
it evaluates the default and, when needed, additional candidates on
$\hat\map$ to select the executed plan (\cref{sec:method-deploy}).

\subsection{The Analytic Referee $\referee$}
\label{sec:method-referee}

Based on the rasterized scene representation employed by the system, the referee module implements the two geometric safety checks corresponding to Drivable Area Compliance (DAC) and No At-Fault Collision (NC). Given a candidate trajectory $\traj$, the current ego speed $v_0$, and the scene representation $\map=(\map_{\mathrm{drv}},\map_{\mathrm{veh}})$, the module executes three deterministic steps.

\noindent\textbf{Kinematic rollout.}
The discrete waypoints in $\traj$ are interpolated and re-simulated in closed loop using the evaluator's linear-quadratic regulator (LQR) tracker and kinematic bicycle model, producing simulated ego poses $\{p_i\}_{i=1}^{N_s}$ at simulation frequency $f_s$. The rollout starts from the origin of the current ego frame at speed $v_0$, which is shared by all candidates of a scene. Here $N_s$ is the number of simulated steps and $p_i=(x_i^s,y_i^s,\psi_i^s)$ is the simulated ego pose at step $i$.
Safety is evaluated on these simulated poses rather than directly on the planned waypoints.

\noindent\textbf{Geometric checks.}
At each $p_i$, we place a fixed set of footprint query points $Q(p_i)$ consisting of the four vehicle corners and center. A drivable-area violation is recorded for pose $i$ if any query point lies outside $\map_{\mathrm{drv}}$. For collision checking, $p_i$ is aligned to the nearest temporal slice of $\map_{\mathrm{veh}}$, and a violation is recorded if any query point lies in an occupied cell.

\noindent\textbf{Metric margins.}
Let $D_{\mathrm{drv}}$ denote the distance transform of the drivable-area map and $D_{\mathrm{veh}}^{(i)}$ that of the vehicle-occupancy slice aligned with $p_i$. Each gives metric clearance to the nearest unsafe cell and is zero on the unsafe side of the corresponding boundary. The referee outputs
\begin{equation}
\setlength{\jot}{1pt}
\begin{aligned}
n_{\mathrm{dac}}
&=\sum\nolimits_{i=1}^{N_s}
\mathbf{1}\!\left[
\exists q\in Q(p_i):\map_{\mathrm{drv}}(q)=0
\right],\\
n_{\mathrm{nc}}
&=\sum\nolimits_{i=1}^{N_s}
\mathbf{1}\!\left[
\exists q\in Q(p_i):\map_{\mathrm{veh}}^{(i)}(q)=1
\right],\\
m_{\mathrm{dac}}
&=\min\nolimits_{i,\;q\in Q(p_i)}D_{\mathrm{drv}}(q),\\
m_{\mathrm{nc}}
&=\min\nolimits_{i,\;q\in Q(p_i)}D_{\mathrm{veh}}^{(i)}(q).
\end{aligned}
\label{eq:referee}
\end{equation}
so that $\referee=(n_{\mathrm{dac}},n_{\mathrm{nc}},
m_{\mathrm{dac}},m_{\mathrm{nc}})$. The two $n$ terms count violating simulated poses, while the two $m$ terms measure minimum clearance to the corresponding hazard. We aggregate them as $n(\traj)=n_{\mathrm{dac}}+n_{\mathrm{nc}}$ and
$m(\traj)=\min(m_{\mathrm{dac}},m_{\mathrm{nc}})$.

The referee contains no learned parameters once the map, trajectory, ego speed, and vehicle geometry are given. Its scope is deliberately limited to these two geometric safety checks: it does not score progress, comfort, or overall plan quality. During training, the referee determines which candidates satisfy the geometric constraints, and its own clearance margin determines preference among those that do.

The rasterized maps and footprint queries used by the referee are not
identical to the privileged geometric representation of the official
evaluator. We therefore measure rather than assume fidelity. Specifically, we audit (i) candidate-level agreement with the evaluator's corresponding safety gates and (ii) the effect of replacing $\map_{\mathrm{gt}}$ with $\hat\map$ on referee verdicts and margins. Both audits are reported in \cref{sec:experiments}.

\subsection{Training-Time Placement: Preference Distillation on Ground-Truth Maps}
\label{sec:method-train}

The base policy $\pi_\theta$ follows the joint video--action generation framework of Cosmos3~\cite{nvidia2026cosmos3omnimodalworld}. Its future
action tokens represent $(x_j,y_j,\cos\psi_j,\sin\psi_j)$ in the ego frame and are trained jointly with the video branch by flow matching. At inference, they are de-normalized and decoded into the trajectory representation defined in \cref{sec:method-overview}.

During preference construction, the referee operates on
$\map_{\mathrm{gt}}$. For each training scene, the policy self-samples $K_t$ trajectories, forming $\mathcal C_t=\{\traj_k\}_{k=1}^{K_t}$. Let $\mathcal P\subset\mathcal C_t$ contain candidates that pass both geometric checks and $\mathcal V\subset\mathcal C_t$ those that violate at least one.
We retain scenes for which both sets are nonempty. Given the
expert trajectory $\traj^\ast$, we construct
\begin{equation}
\setlength{\abovedisplayskip}{4pt}
\setlength{\belowdisplayskip}{4pt}
\traj^w
=
\arg\max_{\traj\in\mathcal P}
m(\traj),
\qquad
\traj^l
=
\arg\min_{\traj\in\mathcal V}
d(\traj,\traj^\ast),
\label{eq:pair}
\end{equation}
where $\traj^w$ and $\traj^l$ are the preferred and dispreferred
trajectories. In short, \emph{safety qualifies, and the referee's own margin decides}: the winner is the passing candidate with the largest safety margin, selected without reference to the expert; the loser is the most expert-like violator, providing a hard negative that remains close in imitation distance while crossing a geometric safety boundary, consistent with prior findings on preference optimization and hard negatives in driving~\cite{shang2026drivedpo,wang2026beyond}. The expert trajectory thus enters pair construction only through the selection of this hard negative.

We continue training with a reference-free pairwise preference objective,
\begin{equation}
\mathcal{L}
=
w_v\,\mathcal{L}^{\mathrm{vid}}_{\mathrm{FM}}
+
w_a\,\mathcal{L}^{\mathrm{act}}_{\mathrm{FM}}(\traj^w)
-
\log\sigma\!\left(
\beta\left[
\ell_\theta(\traj^l)-\ell_\theta(\traj^w)
\right]
\right),
\label{eq:pref}
\end{equation}
where $\mathcal{L}^{\mathrm{vid}}_{\mathrm{FM}}$ and
$\mathcal{L}^{\mathrm{act}}_{\mathrm{FM}}$ are the original video and action flow-matching losses, weighted by $w_v$ and $w_a$;
$\ell_\theta(\traj)$ is the per-sample action flow-matching loss;
$\sigma(\cdot)$ is the logistic function; and $\beta$ controls preference strength. The first two terms retain the base model's flow-matching supervision, the action term on the winner and the video term on the recorded future video, and serve as an anchor without maintaining a frozen reference policy. The pairwise term favors lower action denoising loss for $\traj^w$ than for $\traj^l$. As in diffusion preference optimization~\cite{wallace2024diffusion}, this loss difference is a surrogate for an intractable trajectory log-likelihood ratio rather than an exact likelihood-based DPO objective. Both members of a pair are evaluated at the same flow-matching timestep. Training therefore distills the referee's geometric preference into the policy weights.

\subsection{Deployment-Time Placement: Gated Selection on Self-Predicted Maps}
\label{sec:method-deploy}

At deployment, $\map_{\mathrm{gt}}$ is unavailable and the same referee instead operates on $\hat\map$. The readout $g_\phi$ uses frozen multi-layer visual features from the $H{+}1$ observed frames. Following the LSS/Simple-BEV lineage~\cite{philion2020lift,harley2023simple}, it predicts
per-pixel depth distributions, lifts image features into BEV, and aligns history features using the known ego poses $P_{-H:0}$. Visibility channels distinguish unobserved from observed free space, and a convolutional encoder--decoder predicts a static drivable-area map $\hat\map_{\mathrm{drv}}$ and $1{+}T_f$ time-indexed vehicle-occupancy maps $\hat\map_{\mathrm{veh}}$. During readout training, the visual backbone is frozen and only $\phi$ is optimized, using rasterized privileged annotations from the training set. At deployment, the predicted maps use the same spatial and temporal representation as $\map_{\mathrm{gt}}$, so the referee itself operates unchanged.

Deployment proceeds in two stages. First, a single policy sample
$\bar\traj$ serves as the default plan. From its referee outputs on
$\hat\map$, additional sampling is triggered only when
\begin{equation}
\setlength{\abovedisplayskip}{4pt}
\setlength{\belowdisplayskip}{4pt}
n(\bar\traj)>0
\qquad\text{or}\qquad
m(\bar\traj)\leq\Delta,
\label{eq:deploy-gate}
\end{equation}
where $\Delta$ is the clearance threshold. Otherwise, the default plan is executed directly.

On an alarm, the policy draws $K{-}1$ additional candidates and the referee evaluates all $K$ plans on the same predicted map $\hat\map$. Candidates are ranked first by fewer violating poses, then by larger clearance, and finally by smaller $d(\traj,\bar\traj)$, which only breaks ties. A candidate replaces
$\bar\traj$ only if it reduces the violation count by at least $\delta_n$, or, at equal violation count, improves the clearance by at least $\Delta$; otherwise the default plan is retained. Here $\delta_n$ denotes the required minimum reduction in violating simulated poses. Violation count precedes clearance because the distance transforms are zero after a boundary is crossed, so clearance alone cannot distinguish among violating candidates.

A key design choice is that margins are used comparatively on the
\emph{same predicted map}, rather than as absolute safety certificates. Candidates in one scene therefore share the same map estimate, making their relative ordering less sensitive to common local map bias, although perception error is not eliminated. We measure this error explicitly in \cref{sec:experiments}. The thresholds, candidate counts, and deployment costs are likewise reported there.
\section{EXPERIMENTS}
\label{sec:experiments}

\begin{table*}[!t]
\centering
\caption{Comparison with representative world-model-based planners on
\texttt{navtest} (12,146 scenes). $N{\times}$C denotes $N$ cameras and +L
denotes LiDAR. Video indicates whether video generation is part of the model.
PDMS follows NAVSIM v1 and EPDMS follows v2. NC, DAC, TTC, Comf., and EP
denote no at-fault collisions, drivable-area compliance, time-to-collision,
comfort, and ego progress, respectively. Higher is better; ``---'' denotes
unreported results.}
\label{tab:main}

\setlength{\tabcolsep}{7pt}
\begin{tabular}{llcccccccc}
\toprule
Method & Input & Video &
NC$\uparrow$ & DAC$\uparrow$ & TTC$\uparrow$ &
Comf.$\uparrow$ & EP$\uparrow$ &
PDMS$\uparrow$ & EPDMS$\uparrow$ \\
\midrule

DrivingGPT~\cite{chen2025drivinggpt}
& 1$\times$C & \checkmark
& 98.9 & 90.7 & 94.9 & 95.6 & 79.7 & 82.4 & --- \\

WoTE~\cite{li2025end}
& 3$\times$C+L & \ding{55}
& 98.5 & 96.8 & 94.9 & 99.9 & 81.9 & 88.3 & --- \\

DriveVLA-W0~\cite{li2026drivevla}
& 1$\times$C & \checkmark$^{\dagger}$
& 98.7 & 99.1 & 95.3 & 99.3 & 83.3 & 90.2 & 86.1 \\

PWM~\cite{zhao2026forecasting}
& 1$\times$C & \checkmark
& 98.6 & 95.9 & 95.4 & 100.0 & 81.8 & 88.1 & --- \\

DriveLaW$^{\ddagger}$~\cite{xia2026drivelaw}
& 1$\times$C & \checkmark$^{\dagger}$
& 99.0 & 97.1 & 96.7 & 100.0 & 81.3 & 89.1 & --- \\

CoPhy$^{\S}$~\cite{wu2026distill}
& C+L & \ding{55}
& 99.0 & 98.2 & 96.8 & 100.0 & 85.3 & 91.4 & 86.1 \\

DriveDreamer-Policy~\cite{zhou2026drivedreamer}
& 3$\times$C & \checkmark
& 98.4 & 97.1 & 95.1 & 100.0 & 83.5 & 89.2 & 88.7 \\

Metis~\cite{li2026metis}
& 1$\times$C & \checkmark
& 98.3 & 97.1 & 94.7 & 100.0 & 83.4 & 89.1 & 89.5 \\

DriveVA$^{\ddagger}$~\cite{liu2026driveva}
& 1$\times$C & \checkmark
& 99.2 & 97.5 & 98.7 & 100.0 & 83.5 & 90.9 & --- \\

UNIVERSE~\cite{liu2026universeunifiedvideoaction}
& 1$\times$C & \checkmark$^{\dagger}$
& 99.1 & 97.6 & 98.5 & 100.0 & 83.6 & 91.0 & --- \\

\midrule

Ours (base, 4B backbone)
& 1$\times$C & \checkmark
& 99.22 & 97.37 & 96.86 & 100.00 & 84.09 & 90.47 & 90.03 \\

Ours (full, 4B backbone)
& 1$\times$C & \checkmark
& 99.21 & 98.25 & 96.86 & 100.00 & 85.36 & 91.50 & 91.08 \\

\addlinespace[2pt]
Ours (base, 16B backbone)
& 1$\times$C & \checkmark
& 99.31 & 97.79 & 97.41 & 100.00 & 84.45 & 91.08 & 90.69 \\

\textbf{Ours (full, 16B backbone)}
& 1$\times$C & \checkmark
& \textbf{99.49} & \textbf{98.64} & \textbf{97.91} & \textbf{100.00} & \textbf{84.99}
& \textbf{92.02} & \textbf{91.56} \\

\bottomrule
\end{tabular}

\vspace{2pt}
\raggedright\footnotesize
$^{\dagger}$Trained with video generation but no video is generated at test time.
$^{\ddagger}$DriveVA uses extra CARLA~\cite{dosovitskiy2017carla} data; DriveLaW uses nuPlan~\cite{caesar2021nuplan} and nuScenes~\cite{caesar2020nuscenes}.
$^{\S}$CoPhy uses external VQA data and multi-candidate inference.
Metis reports single-pass inference; its oracle best-of-6 result is excluded.
``Full'' denotes referee distillation with $K{=}2$ gated selection.
Our 4B/16B models use Cosmos3-Edge/Cosmos3-Nano, with 2B/8B trainable generation towers, respectively; the reasoner towers are frozen.
\vspace{-15pt}
\end{table*}

\subsection{Setup}
\label{sec:setup}

We evaluate on the full \texttt{navtest} split of
NAVSIM~\cite{dauner2024navsim}, comprising 12,146 scenes, and report the
official PDMS and EPDMS metrics. Our evaluations use the official one-stage
\texttt{navtest} setting with non-reactive log-replay traffic agents.
PDMS follows NAVSIM v1, while EPDMS uses the NAVSIM v2 extended metric set
on the same split. Both metrics include hard geometric safety gates for
collision and drivable-area compliance.

All controlled comparisons are paired at the scene level on the same base
policy and candidate samples. We report mean score differences with bootstrap
95\% confidence intervals over 10,000 resamples. Internal controlled
comparisons use EPDMS unless otherwise specified.

Our base policy uses Cosmos3-Nano (16B total
parameters)~\cite{nvidia2026cosmos3omnimodalworld}. We fine-tune
its 8B-parameter generation tower on \texttt{navtrain}, while keeping the reasoner tower frozen.
It takes four history frames and the current front-camera frame as visual
conditioning and jointly generates eight future video frames and eight future
action steps. Training, map construction, and referee implementation details
are summarized in \cref{sec:impl}.

\subsection{Referee Fidelity Audits}
\label{sec:calib}

We audit two sources of approximation separately: the referee's agreement
with the official evaluator given ground-truth scene information, and the
additional error introduced when the ground-truth map is replaced by the
self-predicted map used at deployment.

\noindent\textbf{Referee--evaluator fidelity.}
We compare the referee with the corresponding official DAC and collision
gates on trajectories self-sampled by the policy. Because violations are
rare on the full split, we evaluate this audit on the 1,060 \texttt{navtest}
scenes where the base policy scores zero, that is, fails at least one hard
safety gate, sampling five trajectories per scene for a total of
5,300 candidates. Across the resulting gate decisions, the referee achieves
91.0\% agreement with the official evaluator, compared with 73.5\% for an
always-safe predictor. The drivable-area gate attains 0.995 precision and
0.631 recall, while the collision gate attains 0.931 precision and 0.792
recall. Residual disagreement partly reflects the representation mismatch:
the official evaluator uses privileged geometric annotations, whereas our
deployable referee queries vehicle-footprint points on a 0.4\,m occupancy
grid. The high precision supports reliable hard-negative mining, while the
referee's clearance margin selects the preferred passing candidate
(\cref{sec:method-train}).

\noindent\textbf{Map-source fidelity.}
We next isolate the perception error faced at deployment by applying the same
referee to $\map_{\mathrm{gt}}$ and to the readout prediction $\hat\map$ for
the default plan of every \texttt{navtest} scene. The resulting
drivable-area clearance margins $m_{\mathrm{dac}}$ correlate at 0.895, with a
mean absolute difference of 0.502\,m. This audit separates error in
the scene representation from error in the deterministic verdict computation,
and directly measures the uncertainty introduced by replacing privileged maps
with self-predicted ones.

The deployment operating point is fixed before evaluation on
\texttt{navtest}. We use $\Delta=0.4$\,m, corresponding to one occupancy-grid
cell, and $K=2$ candidates for compute-matched comparisons; operating-point
checks are performed only on held-out driving logs disjoint from
\texttt{navtest}.

\subsection{Comparison with Prior Work}
\label{sec:comparison}

\Cref{tab:main} compares \method{} with representative generative
world-action models and end-to-end planners on \texttt{navtest}. With
single-camera visual input and no external training data, the complete system
achieves 92.02 PDMS and 91.56 EPDMS, outperforming
the listed generative world-action models on both reported official metrics;
the distilled policy alone reaches 91.96 PDMS and 91.61 EPDMS in a single
pass. The imitation-only base scores 91.08 PDMS and
90.69 EPDMS; the source of the improvement is analyzed through
controlled comparisons in the following sections.

The gain does not come from policy scale. Repeating the whole pipeline with
the compact Cosmos3-Edge backbone, leaving the readout, referee, and
deployment settings unchanged, improves PDMS from 90.47 to 91.50 and EPDMS
from 90.03 to 91.08, a gain of $+1.03$ PDMS against $+0.94$ for our main
backbone, with gains again concentrated in drivable-area compliance and
ego progress. The compact full system also exceeds the main 16B base
(91.50 versus 91.08 PDMS), showing that the gain is not specific to the
larger backbone.

\subsection{Must the Verdict Be Learned?}
\label{sec:verdict}

\noindent\textbf{Controlled verdict-source comparison.}
We compare different verdict sources while keeping the planning setup fixed.
All methods use the same undistilled base policy, the same pre-sampled
$K{=}2$ candidate trajectories for every \texttt{navtest} scene, and the same
two-stage selection protocol. Only the verdict source changes; each method
uses its own scores to decide when to trigger and whether to replace the
default plan.

We consider two groups of alternatives. The first contains three map-blind
controls (random, smoother, and more conservative) and learned verifiers that
operate directly on visual features, following Hydra-MDP-style metric
distillation~\cite{li2024hydra,li2025hydra} and recent test-time verifier
designs~\cite{tan2026sparsedrivev2,he2026drivever,li2025end,xu2026test}.
Their operating points are selected on held-out driving logs disjoint from
\texttt{navtest}. The second is an additional same-map control, where a learned replacement
receives the same predicted map, candidate trajectory, and motion state as
the analytic referee; we analyze this setting below.

\begin{figure}[t]
\centering
\includegraphics[width=0.9\columnwidth]{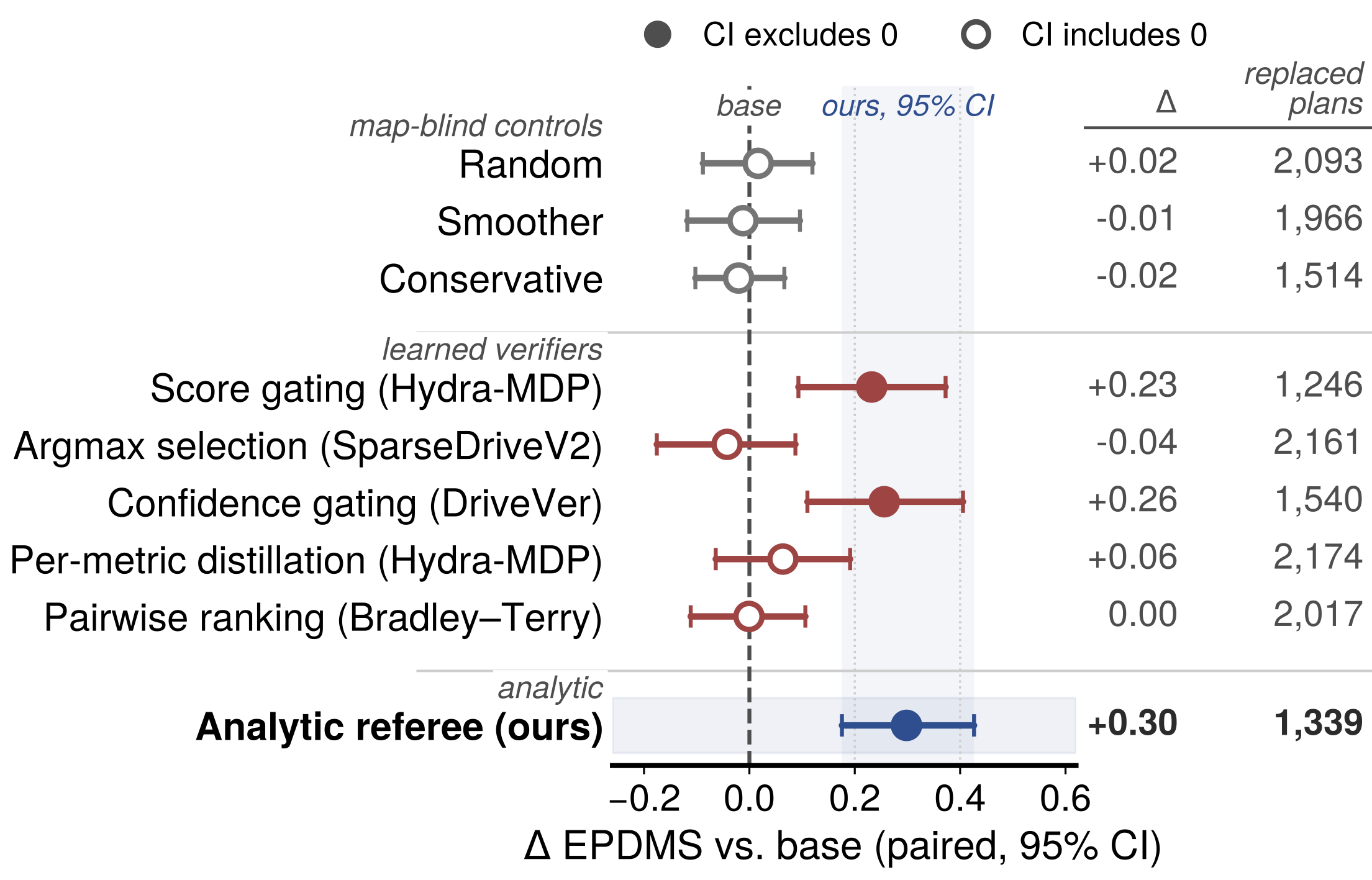}
\caption{\textbf{Matched-budget verdict-source comparison.}
Per-scene paired $\Delta$EPDMS relative to the imitation-only base on
\texttt{navtest} ($n{=}12{,}146$). All sources use the same pre-sampled
$K{=}2$ candidates and are calibrated to an average sampling budget of
about 1.35$\times$. Error bars are bootstrap 95\% confidence intervals.
Numbers at right report replaced plans.}
\label{fig:forest}
\vspace{-15pt}
\end{figure}

\Cref{fig:forest} shows the matched-budget results.
The three map-blind controls give almost no improvement. The best learned
verifiers improve EPDMS by $+0.23$ and $+0.26$, while the analytic referee
reaches $+0.30$. None of the learned verifiers performs significantly better
than the analytic referee under the same sampling budget.

\noindent\textbf{Same-map verdict comparison.}
The comparisons above cover the main learned and heuristic alternatives.
We add a same-map control to isolate the final verdict step by giving the
learned verifier the same predicted map, candidate trajectory, and motion
state as the analytic referee.
For this control, we use a 4.15M-parameter learned verifier based on the
same multi-target distillation design used above.
It receives the same $\hat\map$, candidate trajectory, and motion state as
the analytic referee.
With 66{,}385 evaluator-labeled training candidates, the learned verifier
provides no measurable gain over the analytic referee. In the paired
comparison, the difference is $-0.01$ $\Delta$EPDMS
(95\% CI $[-0.15,+0.14]$), and the two methods select the same plan in
83.4\% of scenes. The learned verifier averages 1.57 policy samples per scene, versus about 1.35 for the analytic referee. It also makes 1.4 times as many plan replacements.
A Bradley--Terry variant with the same predicted map and trajectory information gives no improvement over the base ($-0.10$, 95\% CI $[-0.22,+0.02]$).
Thus, when the scene representation is fixed, learning the final verdict
provides no measurable benefit over direct geometric computation.
\begin{figure}[t]
\centering
\includegraphics[width=0.9\columnwidth]{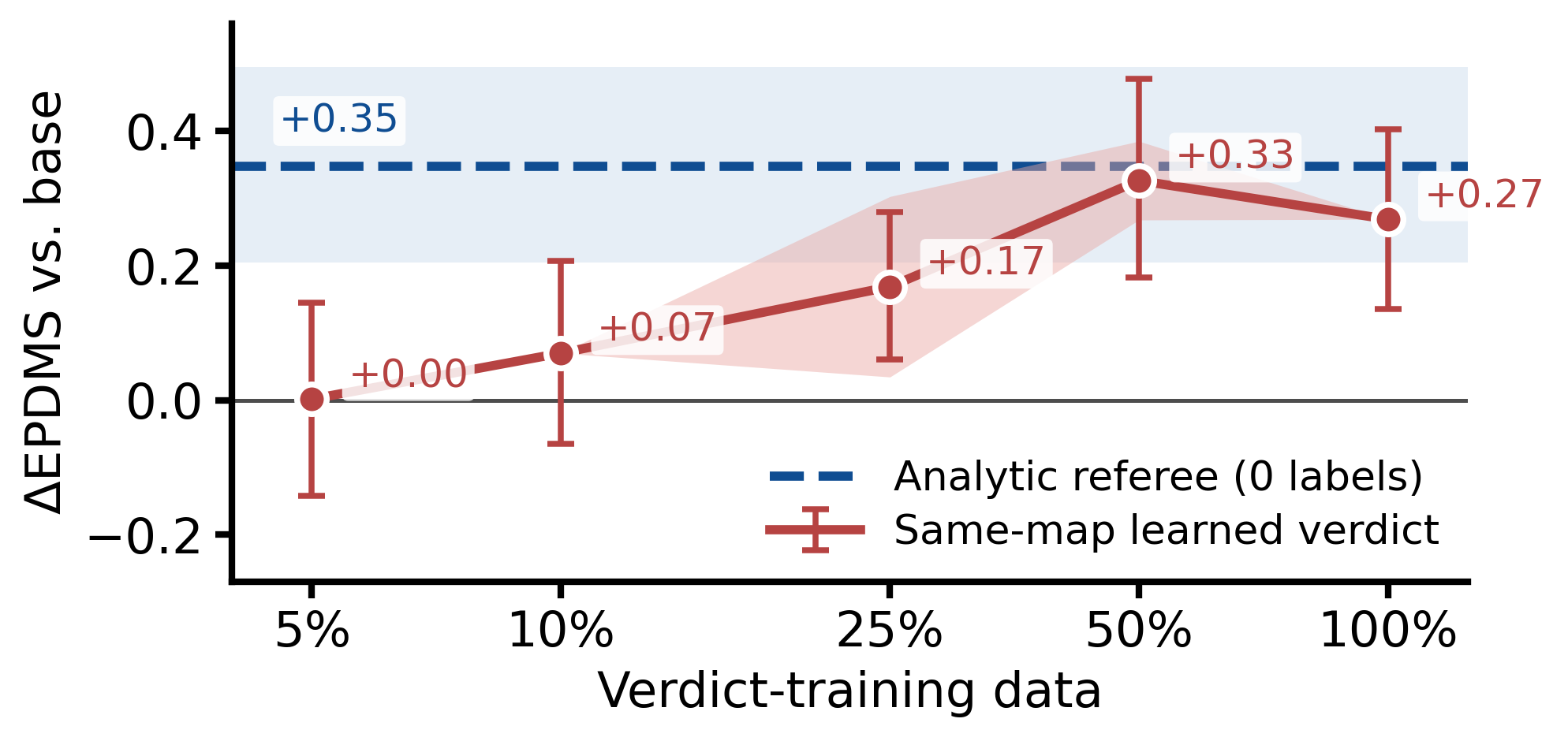}
\caption{\textbf{Verdict-supervision scaling in the same-map setting.}
Only the amount of verdict-training data is varied; 100\% corresponds to
66{,}385 training candidates. Points show the mean over two log-level draws, with bootstrap 95\% confidence intervals. The analytic referee uses no verdict-training labels.}
\label{fig:scale}
\vspace{-15pt}
\end{figure}

\begin{figure*}[t]
\centering
\includegraphics[width=0.95\textwidth]{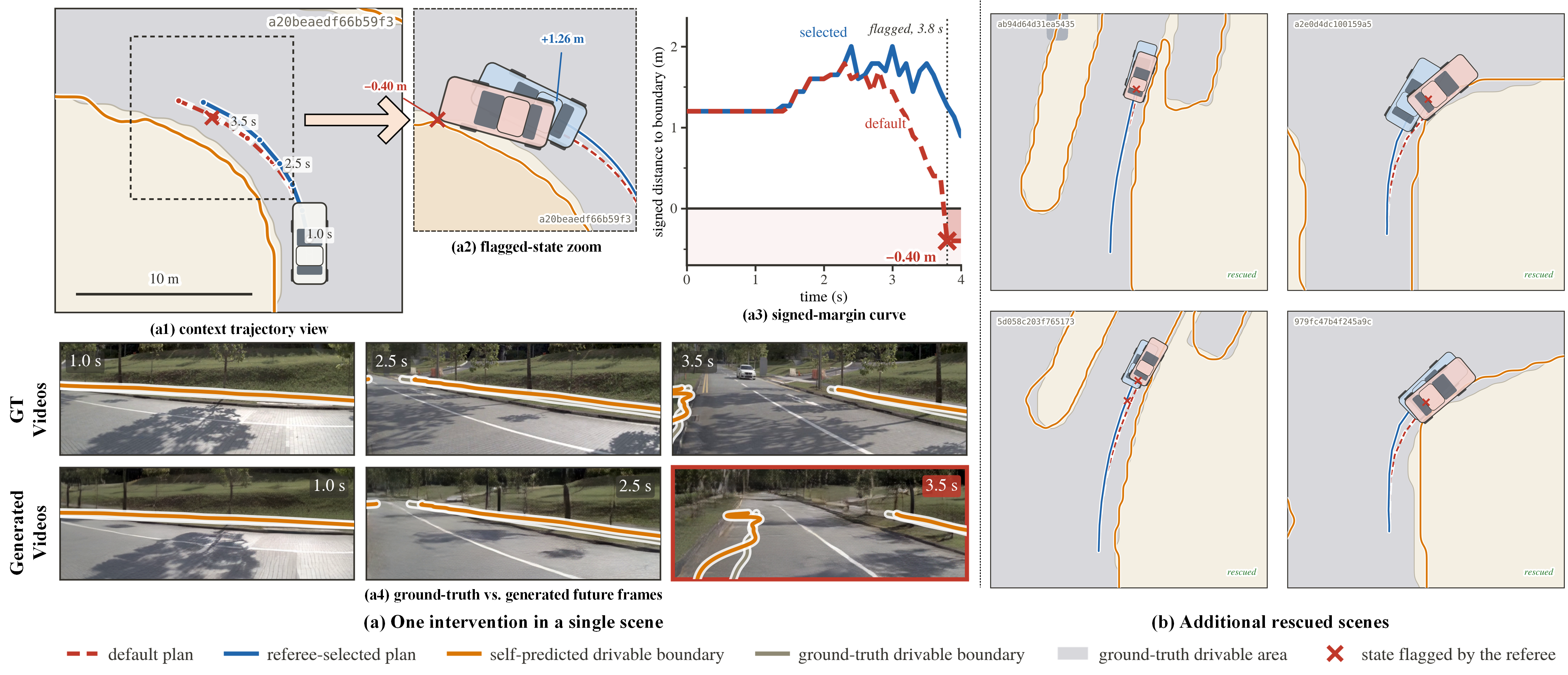}
\caption{\textbf{Deployment-time verification on self-predicted maps}
(base policy, $K{=}2$). 
(a)~One intervention example. At the flagged state, the default plan crosses
the predicted drivable boundary by 0.40\,m, while the referee-selected plan
keeps 1.26\,m clearance. The signed-distance curve and projected boundaries
show the same failure in BEV and video space; the referee's clearance margin
is clipped to zero once the boundary is crossed.
(b)~Four additional replacements found by automatic screening. In each case,
the default plan leaves the ground-truth drivable area, while the selected
plan remains inside both the predicted and ground-truth boundaries
(local IoU $\geq 0.90$).}
\label{fig:qual}
\vspace{-15pt}
\end{figure*}

\noindent\textbf{Supervision scaling.}
Unlike the matched-budget comparison in \cref{fig:forest}, this scaling
study uses a fixed 2.0$\times$ sampling budget for both the learned verifier
and the analytic referee.
We next vary the amount of verdict-training data used by the same-map
learned verifier while keeping its architecture, inputs, candidate set,
decision protocol, and calibration procedure fixed. As shown in
\cref{fig:scale}, using 5\%, 10\%, 25\%, 50\%, and 100\% of the
66{,}385-candidate training split gives $+0.00$, $+0.07$, $+0.17$,
$+0.33$, and $+0.27$ $\Delta$EPDMS, respectively. The analytic referee
reaches $+0.35$ with no verdict-training labels. The learned verifier
approaches this level only after using a large fraction of the available
supervision, and more labels do not give a consistent further gain.

\subsection{One Referee, Two Placements}
\label{sec:decomp}

We evaluate the two uses of the same analytic referee separately.
Training-time preference distillation improves the imitation-only base by
$+0.92$ $\Delta$EPDMS (95\% CI $[+0.71,+1.13]$), while deployment-time
gated selection on the undistilled base gives $+0.30$
($[+0.18,+0.42]$). The full system reaches the highest PDMS in
\cref{tab:main}. After distillation, adding deployment-time selection gives
only $+0.06$ PDMS (95\% CI $[-0.01,+0.14]$) and $-0.04$ EPDMS
($[-0.12,+0.04]$). This shows that most of the benefit of deployment-time
selection is already learned during preference distillation.

We next test whether map prediction limits deployment-time selection. We replace $\hat\map$ with the ground-truth map on the distilled policy while keeping the candidates and decision protocol fixed. With the ground-truth map, deployment-time selection gives $+0.08$ $\Delta$EPDMS (95\% CI $[+0.03,+0.13]$; $+0.10$ PDMS), while the same protocol on $\hat\map$ gives no measurable gain. This suggests that the remaining deployment error mainly comes from map prediction rather than the geometric verdict. Using the same predicted map for all candidates reduces the effect of shared local map errors, but does not remove them.

\Cref{fig:qual} shows representative deployment-time interventions on \texttt{navtest}. On the undistilled base policy, selection fixes an official hard safety-gate failure in 53 scenes and introduces one in 9. Most failures occur when the self-predicted geometry disagrees with the scene geometry used by the official evaluator. These cases agree with the map-source audit in \cref{sec:calib}: once the scene representation is available, the verdict is deterministic, while the main remaining deployment errors come from the predicted map.

\subsection{Ablations}
\label{sec:ablation}

\Cref{tab:ablation} evaluates two design choices in referee-guided
preference training.

\noindent\textbf{Preference objective.}
Removing the pairwise term ($\beta=0$) while keeping the winner trajectories
and all other training settings fixed reduces $\Delta$EPDMS from $+0.92$
to $+0.56$. This shows that training on referee-selected winners alone does
not explain the full gain.

\noindent\textbf{Winner construction.}
Among candidates that pass the referee, choosing the one with the largest
safety margin gives $+0.92$ $\Delta$EPDMS, compared with $+0.63$ when choosing
the candidate closest to the expert (paired difference $+0.29$, 95\% CI
$[+0.14,+0.45]$). Thus, the referee's own safety margin is enough to choose
the winner, and we use the margin-based construction in
\cref{sec:method-train}.
\vspace{-5pt}
\begin{table}[h]
\centering
\caption{\textbf{Training design ablations.} Each block changes one design
choice while keeping the remaining settings fixed. Values are paired
$\Delta$EPDMS over the imitation-only base on \texttt{navtest}
($n{=}12{,}146$).}
\label{tab:ablation}
\setlength{\tabcolsep}{9pt}
\vspace{-5pt}
\begin{tabular}{lr}
\toprule
Variant & $\Delta$EPDMS \\
\midrule
\multicolumn{2}{l}{\textbf{\textit{Preference objective}}} \\
\quad $\beta{=}0$ & $+0.56$ \\
\quad full preference objective & $\mathbf{+0.92}$ \\
\multicolumn{2}{l}{\textbf{\textit{Winner construction}}} \\
\quad nearest to expert & $+0.63$ \\
\quad largest safety margin (ours) & $\mathbf{+0.92}$ \\
\bottomrule
\end{tabular}
\end{table}

\subsection{Video Generation}
\label{sec:video}

As shown in \cref{tab:video}, referee distillation changes FVD from 23.90 to
24.25, LPIPS from 0.4166 to 0.4126, and PSNR from 18.75 to 18.92, showing
no material loss in video-generation quality. Prior-work FVD values are listed
only for context because their evaluation settings differ.

\begin{table}[t]
\centering
\caption{\textbf{Video generation quality.}
Our rows use the same \texttt{navtest} scenes, protocol, and sampling seeds.
Prior-work FVD values are for context only because evaluation settings differ.}
\label{tab:video}
\setlength{\tabcolsep}{2pt}
\vspace{-5pt}
\begin{tabular}{lccccc}
\toprule
Method & FVD$\downarrow$ & LPIPS$\downarrow$ & PSNR$\uparrow$
& Frames & Resolution \\
\midrule
Base (imitation only)
& 23.90
& 0.4166
& 18.75
& 9 @ 2Hz & 832$\times$480 \\

+ referee distillation
& 24.25
& 0.4126
& 18.92
& 9 @ 2Hz & 832$\times$480 \\

\midrule
DriveDreamer-Policy~\cite{zhou2026drivedreamer}
& 53.59 & --- & --- & 9 & 144$\times$256 \\

PWM~\cite{zhao2026forecasting}
& 85.95 & --- & --- & 10 & 128$\times$224 \\

ForgeDrive~\cite{zhong2026forgedrive}
& 69.2 & --- & --- & 8 @ 2Hz & 512$\times$1024 \\

DrivingGPT~\cite{chen2025drivinggpt}
& 142.6 & --- & --- & 12 & 288$\times$512 \\
\bottomrule
\end{tabular}
\vspace{-19pt}
\end{table}

\subsection{Implementation Details}
\label{sec:impl}

Our main policy uses Cosmos3-Nano (16B), with optimization and sampling following its original recipe~\cite{nvidia2026cosmos3omnimodalworld}. The 4B control uses Cosmos3-Edge with the remaining system settings unchanged. The learned-verifier baselines in \cref{fig:forest} use the same base policy, candidate set, supervision pool, and matched sampling budget.
Samples use 832$\times$480 video at 2\,Hz, with four history frames, the
current frame, and eight future frames. Actions are normalized using
training-set quantiles. Preference construction uses $K_t{=}5$ self-samples
per scene. Preference training runs for 850 steps over 462 pairs with learning rate $2{\times}10^{-5}$,
$\beta{=}10$, and $w_v{=}w_a{=}10$.
The BEV readout has 21.23M trainable parameters and is trained on
ground-truth raster maps from privileged annotations. Training and validation
data are split by complete driving logs. The referee uses a 0.4\,m grid,
the evaluator's vehicle dynamics and simulation settings, and five footprint
query points: the four corners and center.
At deployment, we use $\Delta=0.4$\,m, $\delta_n=1$, and $K=2$.
The analytic referee runs on CPU and takes a median of 30\,ms per candidate.
Because a second candidate is generated only after an alarm, the full system
uses 1.34$\times$ as many policy samples on average, with alarms in 34.3\%
of scenes. The command is provided through a fixed prompt
template, with no other text input.

\vspace{-5pt}

\section{CONCLUSION}
\label{sec:conclusion}

We study whether geometric safety verdicts in driving world-action models
need to be learned once the scene representation is available. In controlled
comparisons, learned verifiers do not outperform the zero-parameter analytic
referee; even with the same predicted map and trajectory information, learning
the final verdict gives no measurable downstream benefit. The same referee is
used on ground-truth scene state for training-time preference distillation and
on predicted scene state for deployment-time gated selection. DriveReferee
improves PDMS/EPDMS to 92.02/91.56 on \texttt{navtest}
while preserving video generation quality, supporting the
principle: \emph{learn perception, compute the verdict}.

\noindent\textbf{Limitations and future work.}
Our referee currently covers drivable-area compliance and collision avoidance.
Other rules, such as lane direction and traffic-light compliance, require
additional scene semantics. At deployment, errors in the self-predicted maps
can lead to wrong verdicts; better scene prediction, longer temporal context,
or additional sensors may reduce these errors without changing the analytic
rule. Finally, our referee follows NAVSIM's geometric safety logic on a
rasterized scene representation. Whether the same separation between learned
perception and analytic verdicts works with other evaluators, simulators, and
real-world driving systems remains open.

\section*{ACKNOWLEDGMENT}
We thank Jiawei Yang, the members of the PSI Lab at the University of Southern California,
and our colleagues at Toyota Research Institute and Woven by Toyota for helpful discussions and feedback.
This work was supported by Toyota Research Institute (TRI) through the research thrust
``World Models for Next-Generation Autonomous Driving Policy Learning.''
This article solely reflects the opinions and conclusions of its authors and not TRI or any other Toyota entity.

\bibliographystyle{IEEEtran}
\bibliography{refs}

\end{document}